\documentclass[runningheads]{llncs}
\usepackage[T1]{fontenc}
\usepackage{amsmath}
\usepackage{graphicx,verbatim}
\usepackage{amsfonts}
\usepackage{bm}
\usepackage{marvosym} 
\begin{document}
\title{SAM-Sode: Towards Faithful Explanations for Tiny Bacteria Detection
}
%

\author{Wanying Tan\inst{1} \and
Shuo Yan\inst{2} \and
Dazhi Huang\inst{3} \and
Yazheng Liu\inst{3} \and
Zili Shao\inst{3} \and
Rufeng Chen\inst{3} \and
Hechang Chen\inst{4} \and
Mude Shi\inst{5} \and
Tianxing Ji\inst{2}\textsuperscript{(\Letter)} \and
Sihong Xie\inst{3}\textsuperscript{(\Letter)}}  
\authorrunning{W. Tan et al.}
\institute{Shenzhen University, Shenzhen, China \and
The Second Affiliated Hospital, Guangzhou Medical University, Guangzhou, China\\
\email{jitianxing7021@163.com} \and
The Hong Kong University of Science and Technology (Guangzhou), Guangzhou, China\\
\email{sihongxie@hkust-gz.edu.cn} \and 
Jilin University, Changchun, China \and
Guangdong ACXEL Micro \& Nano Tech Co., Ltd., Guangzhou, China}
  
\maketitle              
\begin{abstract}
Interpretability in object detection provides crucial confidence support for clinical auxiliary diagnosis. However, in tiny bacteria detection, traditional explanation methods often suffer from blurred foreground boundaries and diffuse feature attribution due to the extreme sparsity of target morphological features and severe interference from complex backgrounds. Such limitations hinder the provision of logically coherent morphological evidence. To bridge this gap, we propose a novel eXplainable AI (XAI) framework, \textbf{SAM-Sode}. The framework innovatively transforms initial feature attribution maps into geometry-aware prompts, leveraging the prior knowledge of the foundation model (SAM3) to achieve spatial refinement and morphological reconstruction of the explanatory mappings. Furthermore, we introduce a dual-constraint mechanism based on physical significance and geometric alignment to perform instance-level denoising, generating coherent explanations that better align with human expert intuition. Experimental results on our self-constructed bacteria dataset with complex circuit backgrounds (containing 2,524 images) and other public datasets demonstrate that the proposed method effectively suppresses background redundancy and significantly enhances the decision-making transparency of tiny object detection. 


\end{abstract}

\section{Introduction}
Bacterial infectious diseases have always been a major threat to global public health~\cite{ref_health}. Rapid and accurate bacterial identification is a core prerequisite for clinical precise medication, effective infection control, and antibiotic screening. However, the traditional microbial counting method~\cite{ref_perez1987,ref_otsu1975,ref_gonzalez2004,ref_levner2007} is not only time-consuming but also highly dependent on subjective judgment by humans, which is difficult to meet the immediate and fast requirements of modern clinical medicine. In recent years, the small target detection technology in computer vision has gradually penetrated into the medical imaging field. However, due to the "black box" nature of deep learning models, the prediction results often lack traceable decision logic and cannot provide reliable reference indicators for clinical practice.

To enhance the transparency of complex models, the AI field has introduced interpretability (XAI) techniques. Although the perturbation method~\cite{D-RISE,D-CLOSE},and activation-based method~\cite{SSGrad-CAM++,G-CAME} have achieved certain progress in general scenarios, these posterior or intrinsic interpretation methods generally perform poorly on small targets such as bacteria. Due to the extremely small scale of the targets and significant interference from complex backgrounds—such as intricate circuitry and biological impurities—traditional methods suffer from severe attribution diffusion and foreground-background semantic ambiguity. Specifically, gradient-based attribution maps often exhibit fragmented, point-like distributions, failing to construct logically coherent visual evidence. Conversely, activation-based methods suffer from extensive attribution blurring due to the upsampling mechanism, leading to significant attribution overflow that precludes the precise localization of tiny target boundaries. Currently, custom interpretability frameworks specifically designed for extremely small targets are still scarce.

To address the aforementioned limitations, this paper introduces \textbf{SAM-Sode}, an end-to-end framework that integrates bacteria detection with interpretable analysis. The framework innovatively transforms feature attribution maps into visual prompts for \textbf{SAM3}, achieving spatial enhancement and morphological reconstruction of the foundational explanatory mappings. Furthermore, by incorporating a dual-constraint mechanism involving physical significance and geometric alignment, we propose a robust architecture to refine scattered attributions into coherent morphological evidence. The main contributions of this paper are summarized as follows:
\begin{enumerate}
    \item \textbf{Bacteria Dataset with Complex Circuit Backgrounds:} We construct a "Bacteria Detection Dataset under Complex Circuit Backgrounds" containing 2,524 images, providing a specialized benchmark for validating the interpretability enhancement of SAM-Sode in tiny object detection.
    \item \textbf{First SAM-based Explanatory Refinement Framework:} We propose the first end-to-end architecture that leverages the \textbf{SAM3} foundation model for interpretability refinement, addressing the challenges of fragmented and diffuse attribution responses in tiny object detection.
    \item \textbf{Satisfactory Performance Validation:} Extensive experiments conducted on our self-constructed dataset and other public bacteria datasets demonstrate that SAM-Sode effectively suppresses background redundancy and significantly enhances decision-making transparency.
\end{enumerate}

\section{Method}
Our Pipeline consists of three stages: detection of small objects (Section 2.1), interpretable attribution (Section 2.2), and attribution refinement via mask-guided Constraints (Section 2.4). Firstly, the SR-TOD model is utilized to achieve accurate detection of microscopic bacteria; subsequently, the detection results are input into the interpretability module to extract key attribute attribution points; finally, the detection results and attribution information are input into the interpretability enhancement and evaluation module to obtain the optimal interpretation effect. Fig.~\ref{fig1} illustrates the overall structure of SAM-Sode.

\subsection{Small Object Detection}
\begin{figure}[t]
\includegraphics[width=\textwidth]{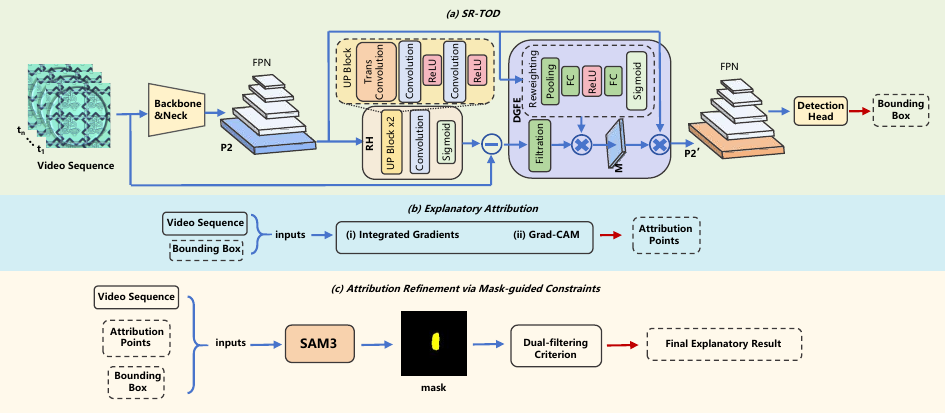}
\caption{The SAM-Sode pipeline: (a) SR-TOD for detection; (b) Explanatory Attribution for point extraction; (c) Attribution Refinement using SAM3 and dual-filtering for final results.} \label{fig1}
\end{figure}

The main difficulty in detecting tiny objects lies in extracting their effective features. SR-TOD enhances the features of tiny objects by constructing the difference graph of the $P2$ layer in the FPN. The structure of this detection module is shown in Fig.~\ref{fig1}.a.

We set the original image as $I_o$. The detection result set $\mathcal{R}$ of the $\text{SR-TOD}$ model (denoted as $F$) is obtained as follows:
\begin{equation}
R = F(I_o), \quad R = \{r_1, r_2, \dots, r_n\}
\end{equation}

\subsection{Explanatory Attribution}

\textbf{Integrated Gradients (IG)}accomplishes feature attribution for target detection concepts from a baseline image to an input image for detection in the form of integration. This method calculates the gradient of the confidence score of the target detection box with respect to the pixels of the input image along the interpolation path, which is used to accumulate the full-process gradient sensitivity of pixels from the baseline image to the actual input image. Finally, the contribution degree is multiplied by the pixel difference weight to explain the contribution of each pixel within the detection box to the model's target detection decision. Specifically, its calculation formula is as follows:

\begin{equation}
IG_i(I_o) = \left(I_{o,i} - I'_{o,i}\right) \times \int_{\alpha=0}^{1} \frac{\partial S\left(F\left(I'_o + \alpha \left(I_o - I'_o\right), r\right)\right)}{\partial I_{o,i}} d\alpha, \quad i \in \Omega_r, \, r \in \mathcal{R}
\end{equation}

We define $\Omega_r$ as the coverage area of a specific instance $r$, and $I'_o$ as the baseline image. \(S(\cdot, r)\) is capable of computing the confidence score of the detection box \(r\).

\textbf{Grad-CAM} processes the top layer of the roi-head in the SR-TOD model, focusing on the gradients corresponding to a specific detection box \( r \). By performing Global Average Pooling (GAP) on the gradients within the range of this detection box, the importance weights of each feature map are obtained. This mechanism can simultaneously capture high-level semantic information and retain its inherent spatial information, thereby explaining the importance of features within the specific detection box in each feature map.

Its specific formula is as follows:
\[
L_{\text{Grad-CAM}}^{c,r} = \text{ReLU} \left( \sum_k \alpha_k^{c,r} \cdot A^k \big|_{\Omega_r} \right)
\]

Among them, the calculation method of the feature map importance weight \( \alpha_k^{c,r} \) for a specific detection box \( r \) is as follows:
\[
\alpha_k^{c,r} = \frac{1}{Z_r} \sum_{(i,j) \in \Omega_r} \frac{\partial S(F(I_o, r), r)}{\partial A_{ij}^k}
\]

 \( Z_r \) is the total number of feature map pixels covered by the detection box \( r \), \( A^k \big|_{\Omega_r} \) represents the part of the \( k \)-th convolutional feature map within the detection box \( r \), and \( S(F(I_o, r), r) \) computes the confidence score of the detection box \( r \) corresponding to the target class \( c \).

\subsection{Segment Anything Model 3 (SAM 3)}
We employ SAM 3 to perform Promptable Concept Segmentation (PCS) \cite{ref_carion2025}, utilizing its multimodal prompting to bridge generic object segmentation and open-set recognition. By integrating text and visual cues, SAM 3 refines instance masks across frames, ensuring temporal consistency and morphological precision for tiny bacterial detection.

\subsection{Attribution Refinement via Mask-guided Constraints}
\subsubsection{Random Slicing Strategy}

To transform discrete attribution distributions into coherent semantic masks, we utilize each individual attribution point as an independent point prompt. These prompts are fed into the Segment Anything Model 3 (SAM3) alongside the detector's predicted bounding box, $B_{pred}$. The inclusion of $B_{pred}$ serves as an explicit spatial prior, effectively suppressing background noise inherent in tiny object detection and ensuring that the generated masks are strictly aligned with the detector's decision regions.

\begin{figure}
\centering
\includegraphics[width=0.8\textwidth]{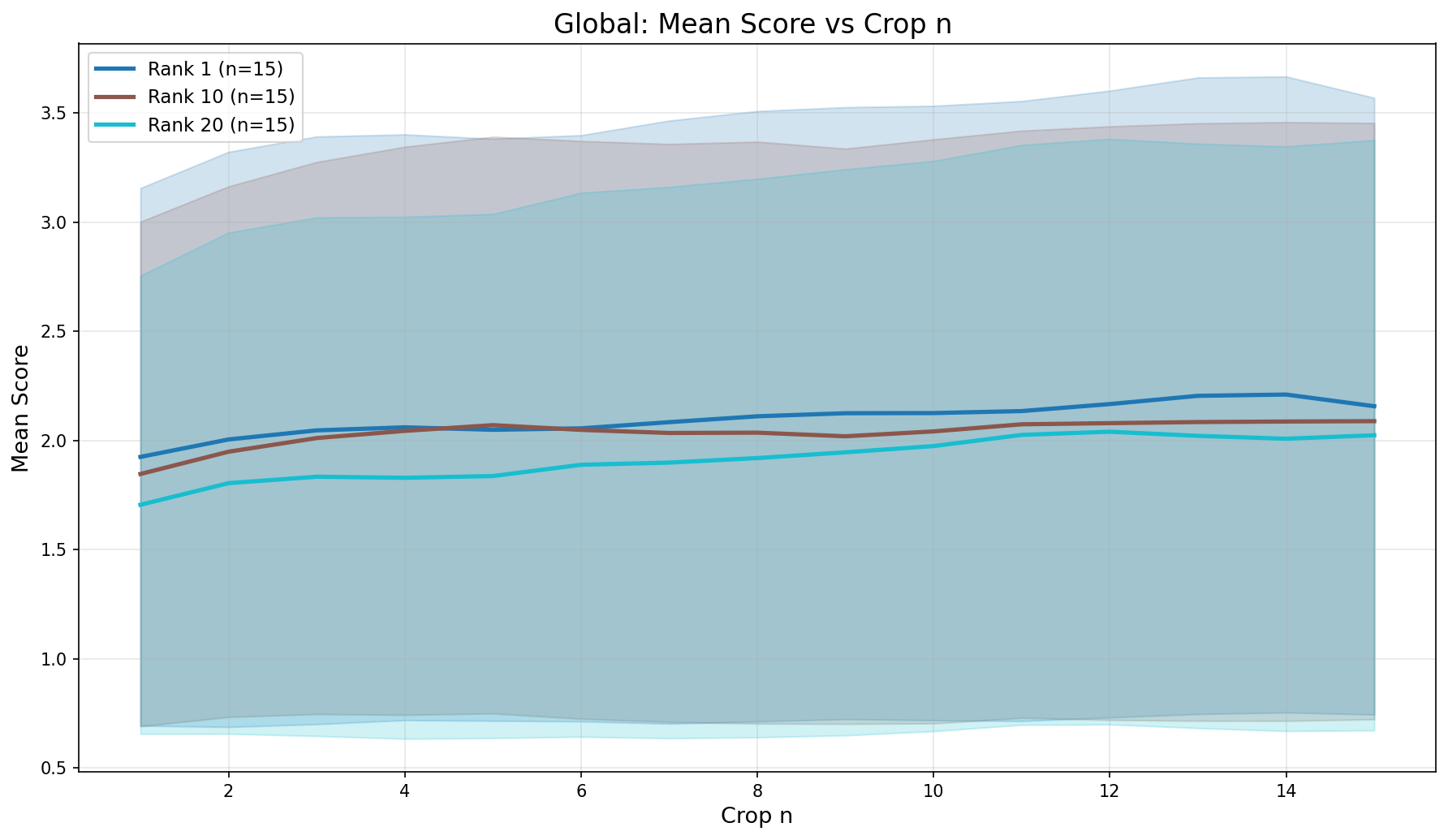}
\caption{\textbf{Global sensitivity analysis of augmentation crops ($n$)}. Mean confidence scores are reported across Rank 1, 10, and 20 (shading denotes $\pm1\sigma$). A performance plateau is observed starting from $n=10$, justifying its selection as the default hyperparameter to balance attribution robustness and computational efficiency.} 
\label{fig3}
\end{figure}

To mitigate the inherent stochasticity of mask generation and enhance structural stability, we propose a \textbf{Random Slicing Strategy}. For each target, we randomly crop $n$ contextual regions centered around $B_{pred}$, with each region's dimensions set to 20 times the area of the target. These cropped regions serve as the diverse image inputs for SAM. The final \textbf{Enhanced Mask (EM)} is defined as the maximum overlapping region (intersection) across these $n$ iterations to ensure morphological consistency. Ablation studies (refer to Fig.~\ref{fig3}) demonstrate that $n=10$ provides the optimal balance between mask robustness and computational efficiency.

\subsubsection{Quantitative Assessment of Enhanced Mask Quality}

To rigorously evaluate the quality of the Enhanced Mask (EM), we introduce evaluation metrics from two dimensions: geometric alignment and physical significance.

\paragraph{Mask Intersection over Union (MaskIoU).}
Given that the ground truth (GT) of tiny objects exists only in the form of bounding boxes, we employ the axis-aligned bounding box of the mask for cross-representation evaluation. Let $S$ be the set of coordinates of all object pixels in the mask:
\begin{equation}
    S = \{ (x, y) \mid M(x, y) = 1 \}
\end{equation}
The boundaries of the bounding box $B_{mask}$ determined by this are:
\begin{equation}
    x_{min} = \min_{(x,y) \in S} x, \quad y_{min} = \min_{(x,y) \in S} y; \quad x_{max} = \max_{(x,y) \in S} x, \quad y_{max} = \max_{(x,y) \in S} y
\end{equation}
Let the ground truth box be $B_{gt}$. The formula for $MaskIoU$ is:
\begin{equation}
    MaskIoU = \frac{|B_{mask} \cap B_{gt}|}{|B_{mask} \cup B_{gt}|} = \frac{\text{Area}(B_{mask} \cap B_{gt})}{\text{Area}(B_{mask}) + \text{Area}(B_{gt}) - \text{Area}(B_{mask} \cap B_{gt})}
\end{equation}

\begin{figure}[t] 
\centering
\includegraphics[width=0.6\textwidth]{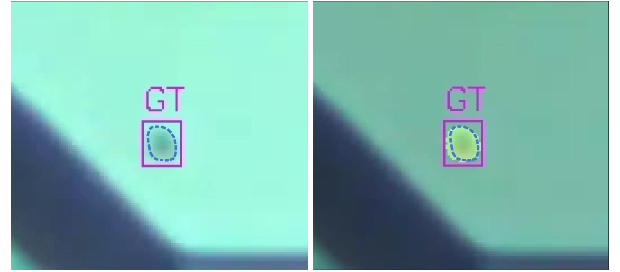}
\caption{(Left) Original bacterial instance; (Right) The corresponding refined mask. The blue dashed lines represent the true physical boundaries of the bacterium.}\label{fig2}
\end{figure}

\paragraph{Mask Physical Significance Score (MaskScore).}
In the scenario of tiny bacteria detection, since $B_{gt}$ often contains the surrounding background (not tightly fitted), blind expansion of the mask within the box may lead to an artificially high $MaskIoU$. Therefore, we refer to the Fisher discriminant criterion and introduce the physical significance index $MaskScore$ to evaluate the precision of the mask:
\begin{equation}
    MaskScore = \frac{(\mu_{core} - \mu_{bg})^2}{\sigma_{core}^2 + \sigma_{bg}^2 + \lambda \cdot \text{Area}_{overflow}}
\end{equation}
Where $core$ is the enhanced mask region, and $bg$ is the background part within the GT box not covered by the mask. $\mu_x$ and $\sigma_x^2$ represent the mean and variance of pixels in the corresponding regions of the original image. $\lambda \cdot \text{Area}_{overflow}$ is the geometric constraint term used to punish the overflow mask that exceeds the GT range, where the penalty factor $\lambda$ is set to the reciprocal of the GT area to achieve adaptive overflow suppression. Fig.~\ref{fig2} shows the relevant areas of one of the enhanced masks.

\subsubsection{Attribution Refinement via Dual Constraints}

To address the background redundancy and fragmented response issues of traditional attribution methods (such as IG and Grad-CAM) in tiny object detection, we propose a refinement mechanism based on spatial consistency. This mechanism does not directly reconstruct the mask but optimizes the original attribution mapping by using the geometric and physical properties of the Enhanced Mask (EM) as filtering criteria:

\paragraph{Dual-filtering Criterion.} For each candidate target, we first extract its corresponding set of attribution points and calculate the $MaskScore$ and $MaskIoU$ for each point. Subsequently, we normalize these metrics and construct a dual-constraint gate: only key explanation units that satisfy $Score_{norm} > 0.4$ and $IoU_{norm} > 0.3$ are retained.

\paragraph{Instance-wise Normalization Isolation.} During processing, we perform independent normalization for each object instance. This isolation mechanism aims to eliminate the inequality of feature intensity between hard and easy samples, ensuring that the most discriminative explanatory features can be extracted under different detection confidence levels.

The refined attribution map provides morphological evidence that better aligns with the intuition of human experts, and the interpretability of the model's decision-making is significantly enhanced. The specific comparison of the enhancement effects will be presented in detail in Section 3.3.

\section{Experiments}
\subsection{Datasets}
The datasets we used include AGAR and a self-built tiny and blurry bacteria dataset with complex circuit backgrounds named TBC-Micro. TBC-Micro consists of 91 video sequences, and by annotating every 3 frames, a total of 2,524 images and 57,472 bounding box annotations were obtained. To ensure the quality of annotations, each bounding box fits the bacterial edges as closely as possible. AGAR provides professional-grade agar plate images; focusing on the core of ``tiny object'' research, we extracted a specific subset containing 1,802 images from it. Following standard experimental procedures, we randomly partitioned the total sample size into training, validation, and test sets at a ratio of 7:1:2.

\subsection{Implementation Details}
To mitigate feature loss for tiny bacterial targets, all input images are processed at their original resolution. In the SR-TOD framework, the scales of the anchor\_generator are set to $[2, 4, 8]$, with corresponding strides of $[4, 8, 16, 32, 64]$. The confidence threshold during inference is established at 0.7. The model is trained for 20 epochs using the SGD optimizer with an initial learning rate of 0.00125. Regarding the Integrated Gradients module, the number of interpolation steps $n_{steps}$ is set to 30.

\subsection{Result}
\begin{figure}[t]
\centering
\includegraphics[width=0.4\textwidth]{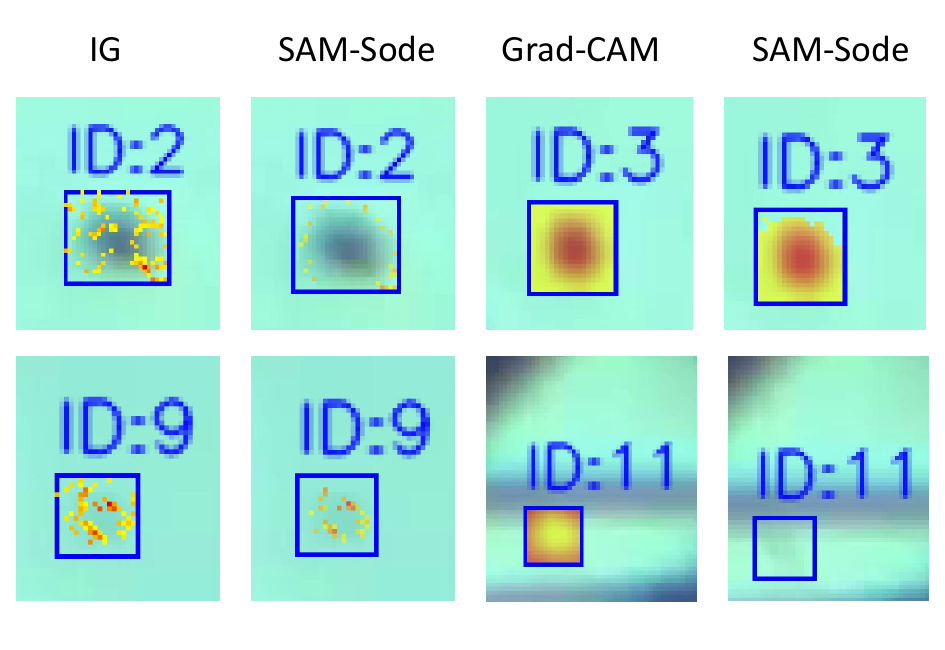}
\caption{For the discrete and fragmented maps generated by IG (e.g., ID:2, 9), our method extracts coherent attribution points that align with bacterial contours. For Grad-CAM results (e.g., ID:3, 11), SAM-Sode purifies background noise and successfully filters out false high-activation signals at the corners, which is particularly evident in the complete suppression of the spurious activation in ID:11.}\label{result}
\end{figure}

As shown in Fig.~\ref{result}, our \text{SAM-Sode} framework significantly enhances the visual quality of explainability and effectively eliminates redundant attributions in the background. For the discrete and fragmented attribution maps generated by \text{Integrated Gradients (IG)}, our method can extract key attribution points that closely fit the morphological contours of bacteria. For \text{Grad-CAM}, where the activation area covers the entire bounding box due to the upsampling mechanism, \text{SAM-Sode} can not only purify part of the background noise but also effectively identify and filter out high-activation false signals at the corners of the bounding box.

\section{Conclusion}
In this paper, we proposed \textbf{SAM-Sode}, a novel approach designed to address the issues of fragmented and diffused attribution maps generated by traditional explainable methods when dealing with tiny objects. By innovatively utilizing attribution salient points as prompt inputs for \textbf{SAM 3} and introducing a \textbf{dual-constraint refinement mechanism}, we achieved a deep optimization of traditional attribution results. Experimental results on our self-collected \textbf{TBC-Micro} dataset (containing 57,472 annotations) and the \textbf{AGAR} petri dish dataset demonstrate that our method effectively eliminates background redundancy and significantly enhances the decision transparency of tiny object detection.

\begin{credits}
\subsubsection{\ackname}We would like to express our gratitude to the scientific researchers of the collaborating hospital for their technical support during the construction of the TBC-Micro dataset. We also thank the company that provided the microfluidic device for collecting bacterial videos. The names of the relevant institutions and personnel have been anonymized in accordance with the requirements of the double-blind review. 

\subsubsection{\discintname}
The authors have no competing interests to declare that are relevant to the content of this article.
\end{credits}

%
%
%

\begin{thebibliography}{22}
\bibitem{ref_health}
Murray, C.J., Ikuta, K.S., Sharara, F., et al.: Global burden of bacterial antimicrobial resistance in 2019: a systematic analysis. \textit{The Lancet} \textbf{399}(10325), 629--655 (2022)

\bibitem{ref_perez1987}
Perez, A., Gonzalez, R.C.: An iterative thresholding algorithm for image segmentation. \textit{IEEE Trans. Pattern Anal. Mach. Intell.} \textbf{9}(6), 742--751 (1987)

\bibitem{ref_otsu1975}
Otsu, N.: A threshold selection method from gray-level histograms. \textit{Automatica} \textbf{11}(3), 23--27 (1975)

\bibitem{ref_gonzalez2004}
Gonzalez, R.C., Woods, R.E., Eddins, S.L.: Digital Image Processing Using MATLAB. Pearson Education India (2004)

\bibitem{ref_levner2007}
Levner, I., Zhang, H.: Classification-driven watershed segmentation. \textit{IEEE Trans. Image Process.} \textbf{16}(5), 1437--1445 (2007)

\bibitem{D-RISE}
Petsiuk, V., Jain, R., Manjunatha, V., Morariu, V.I., Mehra, A., Ordonez, V., Saenko, K.: Black-box explanation of object detectors via saliency maps. In: Proceedings of the IEEE/CVF conference on computer vision and pattern recognition. pp. 11443--11452 (2021).

\bibitem{ref_petsiuk2021}
Petsiuk, V., Jain, R., Manjunatha, V., et al.: Black-box explanation of object detectors via saliency maps. In: Proceedings of the IEEE/CVF Conference on Computer Vision and Pattern Recognition. pp. 11443--11452 (2021)

\bibitem{D-CLOSE}
Truong, V.B., Nguyen, T.T.H., Nguyen, V.T.K., et al.: Towards better explanations for object detection. arXiv preprint arXiv:2306.02744 (2023)

\bibitem{SSGrad-CAM++}
Yamauchi, T.: Spatial sensitive Grad-CAM++: Improved visual explanation for object detectors via weighted combination of gradient map. In: Proceedings of the IEEE/CVF Conference on Computer Vision and Pattern Recognition. pp. 8164--8168 (2024)

\bibitem{G-CAME}
Nguyen, Q.K., Nguyen, T.T.H., et al.: Efficient and concise explanations for object detection with gaussian-class activation mapping explainer. arXiv preprint arXiv:2404.13417 (2024)

\bibitem{LRP}
Montavon, G., Binder, A., Lapuschkin, S., et al.: Layer-wise relevance propagation: An overview. In: Samek, W., Montavon, G., Vedaldi, A., Hansen, L.K., Müller, K.R. (eds.) Explainable AI: Interpreting, Explaining and Visualizing Deep Learning. pp. 193--209. Springer, Cham (2019)

\bibitem{DSSD}
Fu, C.Y., Liu, W., Ranga, A., et al.: DSSD: Deconvolutional single shot detector. arXiv preprint arXiv:1701.06659 (2017). https://doi.org/10.48550/arXiv.1701.06659

\bibitem{SSD}
Liu, W., Anguelov, D., Erhan, D., et al.: SSD: Single shot multibox detector. In: Leibe, B., Matas, J., Sebe, N., Welling, M. (eds.) Computer Vision -- ECCV 2016. pp. 21--37. Springer, Cham (2016)

\bibitem{ION}
Bell, S., Zitnick, C.L., Bala, K., et al.: Inside-outside net: Detecting objects in context with skip pooling and recurrent neural networks. In: Proceedings of the IEEE Conference on Computer Vision and Pattern Recognition. pp. 2874--2883 (2016)

\bibitem{PFFNET}
Chen, B., Solebo, A., Shi, D., et al.: Minuscule cell detection in AS-OCT images with progressive field-of-view focusing. In: Gee, J.C., et al. (eds.) Medical Image Computing and Computer Assisted Intervention -- MICCAI 2025. pp. 365--375. Springer, Cham (2026)

\bibitem{ref_carion2025}
Carion, N., Gustafson, L., Hu, Y.T., Debnath, S., Hu, R., Suris, D., Ryali, C., Alwala, K.V., Khedr, H., Huang, A., et al.: Sam 3: Segment anything with concepts. \textit{arXiv preprint arXiv:2511.16719} (2025)





\end{thebibliography}
%

\end{document}